\pdfoutput=1
\documentclass[11pt]{article}

\usepackage[]{acl}

\usepackage{times}
\usepackage{latexsym}

\usepackage[T1]{fontenc}
\usepackage[utf8]{inputenc}

\usepackage{microtype}
\usepackage{algorithm}
\usepackage{algpseudocode}

\usepackage{caption,subcaption}
\usepackage{amssymb}
\usepackage{graphicx}
\usepackage{amsmath}
\usepackage{multirow, epstopdf}
\usepackage{amssymb}
\title{\begin{flalign*}
    \text{V } & \text{Attack: Taking advantage of Text Classifiers' horizontal vision} \\
        \text{e } & \\
       \text{r } &\\
       \text{t } 
       \end{flalign*}\vspace{-3cm}}
\author{Jonathan Rusert
\\
  Purdue University, Fort Wayne\\
  \texttt{jrusert@pfw.edu }}
\date{June 2022}

\begin{document}

\maketitle

\begin{abstract}
    Text classification systems have  continuously improved in performance over the years. However, nearly all current SOTA classifiers have a similar shortcoming, they process text in a horizontal manner. Vertically written words will not be recognized by a classifier. In contrast, humans are easily able to recognize and read words written both horizontally and vertically. Hence, a human adversary could write problematic words vertically and the meaning would still be preserved to other humans. We simulate such an attack, \textit{VertAttack}. \textit{VertAttack} identifies which words a classifier is reliant on and then rewrites those words vertically. We find that \textit{VertAttack} is able to greatly drop the accuracy of 4 different transformer models on 5 datasets. For example, on the SST2 dataset, \textit{VertAttack} is able to drop RoBERTa's accuracy from 94 to 13\%. Furthermore, since \textit{VertAttack} does not replace the word, meaning is easily preserved. We verify this via a human study and find that crowdworkers are able to correctly label 77\% perturbed texts perturbed, compared to 81\% of the original texts. We believe \textit{VertAttack} offers a look into how humans might circumvent classifiers in the future and thus inspire a look into more robust  algorithms. 
\end{abstract}

\section{Introduction}
\begin{figure}
    \centering
    \begin{subfigure}[b]{0.95\columnwidth}
        \centering
        \fbox{\includegraphics[width=\textwidth]{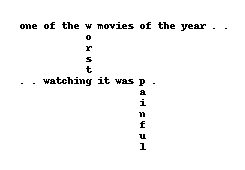}}
    \end{subfigure}
    \begin{subfigure}[b]{0.95\columnwidth}
        \centering
        \fbox{\includegraphics[width=\textwidth]{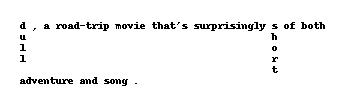}}
    \end{subfigure}
    \caption{Examples of texts perturbed by VertAttack. Humans can still understand the vertically written words, while classifiers struggle to read.}
    \label{fig:examples}
\end{figure}

Automatic text classifiers have seen a continual increase in helping websites moderate and monitor products or people. Though they are helpful to reduce the work load of humans, they can be subject to problems like bias \cite{chuang2021mitigating, zhou2021challenges} and are vulnerable to adversarial attacks \cite{lei-etal-2022-phrase, le-etal-2022-perturbations}. Research into text adversarial attacks has been on the rise in recent years. The reasons range from testing classifiers' robustness \cite{wang-etal-2022-semattack} to privacy concerns \cite{xie-hong-2022-differentially}. 

Current state-of-the-art (SOTA) attacks largely fall into character based attacks and word-based attacks. Character-based attacks change individual characters, by flipping character, introducing or removing whitespace \cite{Grndahl2018AllYN}, or replacing characters with visually similar characters \cite{eger-etal-2019-text}. Word-based attacks replace words with similar words which are less known to the target classifier \cite{li-etal-2020-bert-attack, wang-etal-2022-semattack}. One weakness of current SOTA attacks is that they constrain themselves to horizontal changes. That is, the final result is still read in a left-to-right (English) manner. 
This is a disadvantage because the attacker restricts themselves to the same domain as the classifier which is also only able to read text horizontally. 

%SOTA automated classifiers (and other NLP systems) have a similar shortcoming, they only are able to read text in a horizontal manner. 
Humans have the ability to read text in multiple directions, not just horizontally. Thus, a human attacker who wants to communicate a message to others, while avoiding a website automatically classifying that text, could write the words vertically and the meaning would still be preserved. We simulate this with \textit{VertAttack}. 

%Though less obvious, it is an exploit that can leveraged to hide words in plain sight.
 %Specifically, words can be rewritten vertically and a text classifier will no longer recognize that word.% Unlike automated classifiers, humans are easily able to recognize and read words written both horizontally and vertically. 

\textit{VertAttack} exploits the current limitation of classifiers' inability to read text vertically. Specifically, \textit{VertAttack} perturbs input text by changing information rich words from horizontally to vertically written. Our research makes the following contributions:

1. Propose an attack (\textit{VertAttack}) to mimic how humans may subvert automatic classifiers. This attack exploits current classifiers' glaring weakness (inability to ``read'' vertical text). 

2. Test \textit{VertAttack} on 5 datasets, against 4 different classifiers. We further examine transferability of our attack. We find that when \textit{VertAttack} has blackbox access to the classifier, it is able to drop classification accuracy from 83 - 95\% down to 1 - 36\%. We further compare \textit{VertAttack} with two other text attacks, BERT-ATTACK and Textbugger. We find that, on average, \textit{VertAttack} is able to drop classifiers' accuracy to 36.6\% accuracy, which is lower than BERT-ATTACK (47.5\%) and Textbugger (63.2\%).

3. Verify \textit{VertAttack}'s ability to be understood by humans via qualitative analysis. We find that humans are able to correctly classify 77\% perturbed texts compared to 81\% of the original texts.

4. Investigate initial defenses in terms of whitespace removal and find that if \textit{VertAttack} a classifier reverses the algorithm it is able to mitigate the attack, but simpler whitespace preprocessing is not as effective.

5. Enhance \textit{VertAttack} by allowing it to add in \textit{chaff} to further disguise the text. This chaff greatly affects the reversal defense.  Furthermore, we investigate how \textit{VertAttack} affects classifiers using OCR to extract text from images.

The success of \textit{VertAttack} shows a vulnerability in classifiers which humans may leverage to easily defeat them. We share code and perturbed texts for future research\footnote{We make our code and generated texts available at https://github.com/JonRusert/VertAttack}.

\section{Threat Model}
The examined threat model follows from prior research \cite{formento-etal-2023-using, le-etal-2022-perturbations, deng-etal-2022-valcat}. We assume blackbox knowledge of a classifier. That is, \textit{VertAttack} has no internal knowledge of the classifier, but has access to the probabilities and label output by the model. \textit{VertAttack} uses this for feedback (Section \ref{sect:methods}). 

With prior research, there is an assumption that the feedback classifier is the same as the target classifier. However, websites rarely share the exact classifier used for moderating texts. Thus, we also examine the cases of where the feedback classifier differs from the target classifier as a transferability problem.  

\section{Attack Goals} Based on prior research \cite{lei-etal-2022-phrase, zang-etal-2020-word, Li2019Textbugger} \textit{VertAttack} has 2 goals: 1. Modify text in such a way to cause an automated classifier to fail (misclassify). 2. Ensure modified retains the original meaning to humans. Thus, the attack is similar to obfuscation from classifiers.

Some previous text attack research have made the argument that attacks should be imperceptible to humans \cite{dyrmishi-etal-2023-humans}.  However, this is not a unanimous requirement from text attacks, as many do not include it as a prerequisite \cite{alzantot2018generating, ebrahimi2018hotflip, eger-etal-2019-text, li2021contextualized}. Furthermore, this would disqualify nearly all character-level attacks since humans do not naturally substitute characters in their writing (beyond mispellings). Finally, as stated, \textit{VertAttack} simulates how humans can attack automated classifiers. Thus, we focus on the two aforementioned goals.

\section{Methodology}
\begin{figure*}
    \centering
    \includegraphics[width=\textwidth]{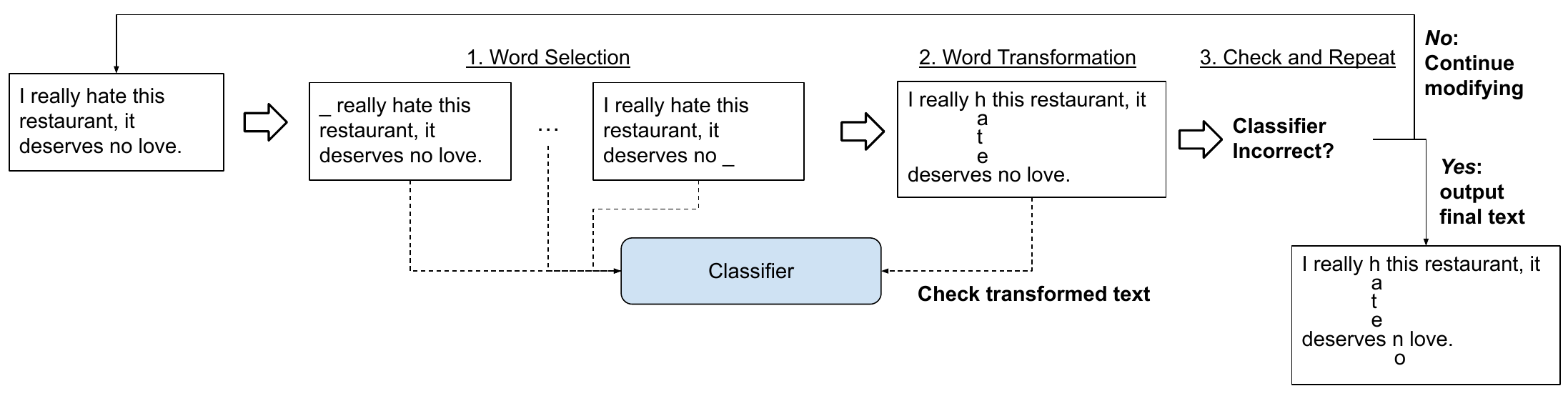}
    \caption{\textit{VertAttack} basic overview. A word to transform is first selected from the input text and then transformed vertically. The classifier assists in providing feedback in the form of class probabilities. The process is repeated until the classifier misclassifies the text.}
    \label{fig:vertattack}
\end{figure*}

Our proposed attack, \textit{VertAttack}, can be divided into two main steps: 1) Word Selection, 2) Word Transformation. A visualization of the method can be seen in Figure \ref{fig:vertattack}. 

\subsection{Word Selection}\label{sect:methods}

\begin{algorithm}
\caption{Word Selection}\label{alg:selection}
\begin{algorithmic}
\Require $text$
\Ensure $j \gets \text{PositionToModify}$
\State $Score_{Orig} \gets \text{Classifier(text)}$
\State $Drop_{Max} \gets 0, i \gets 0, j \gets 0$
\While{$i \ne len(text)$}
    \State $Score_{w} \gets \text{Classifier(text/w)}$
    \State $Drop_{w} \gets Score_{Orig} - Score_{w}$
    \If{$Drop_{w} > Drop_{Max}$}
        \State $Drop_{Max} \gets Drop_{w}$
        \State $j \gets i$
    \EndIf
    \State $i \gets i + 1$
\EndWhile
\end{algorithmic}
\end{algorithm}

First the attack finds which word most helps the classifier. We employ a greedy search method (Algorithm \ref{alg:selection}). In previous work this has been referred to as word importance \cite{jin2020bert} or greedy selection \cite{hsieh-etal-2019-robustness}. The method removes one word\footnote{Here a word is defined as a token separated by whitespace.} at a time and checks the change in classification probability from the original text. Each word is removed and then replaced until all probabilities are calculated. The word that causes the highest drop in probability is chosen as the word to be transformed. 

\subsection{Word Transformation}
\begin{algorithm}
\caption{Word Transformation}\label{alg:transformation}
\begin{algorithmic}
\Require $text, perturb_{positions}$
\Ensure $text_{m}$
\State $\#_{lines} \gets \text{max length of words to be modified}$
\State $k \gets 0$
\While{$k < \#_{lines}$}
    \State $i \gets 0$
    \While{$i \ne len(text)$}
        \If{$i \in perturb_{positions}$}
            %\If{$line < len(text[i])$}
            %    \State $text_{m} \gets text_{m} + text[i][k]$
            %\EndIf
            \State $ \text{append text[i][k]}$
        \Else
            %\If{$k = 0$}
            %     \State $text_{m} \gets text_{m} + text[i]$
            %\Else
            \State $ \text{append word on first line} $
            \State $ \text{or pad spaces equal to word length}$
            %\EndIf
            %\State $text_{m} \gets text_{m} + \text{" "}$
        \EndIf
        \State $\text{append space}$
        \State $i \gets i + 1$
    \EndWhile
    \State $\text{Add newline char to } text_{m}$
    \State $k \gets k + 1$
\EndWhile
\end{algorithmic}
\end{algorithm}

Once a word is selected, it is then transformed vertically (Algorithm \ref{alg:transformation}). First, the number of lines needed (ie. length of word) for each selected word is calculated. Next, we iterate through each word of the original text. If a word is a non-selected word, then it is simply added to the final text. If the word is a selected word, then only the character of the corresponding line is chosen. For example, if ``happy'' is selected, and the line number is 2, then ``a'' is added to the final text. For all lines that only consist of whitespace and the vertical characters, the required whitespace is calculated by the length of each non-selected word. 

Finally, we add a width constraint to the algorithm for practicality. The transformation is only run on that width (number of words) at a time and all text is combined at the end. For example, if there are 100 words and the width constraint is 10, then only 10 are modified at a time. 

Once the transformations are applied, the classifier is queried again to see if the transformed text causes the classifier to misclassify. If so, the final text is produced. If not, then the algorithm repeats, however, this time the words that have been selected already are removed as candidates during the selection step.

\section{Experimental Setup}
To test the effectiveness of \textit{VertAttack}, we evaluate the attack against several transformer classifiers across datasets examined in previous attack papers\footnote{The majority of attacks were run on 56-core 256G processors. \textit{VertAttack} was limited to 1 hour for each attacked text, after 1 hour the attack was noted as failure and no perturbations were made to the text.} \cite{li-etal-2020-bert-attack, jin2020bert, ren-etal-2019-generating, wang-etal-2022-semattack}. 

%\noindent \textbf{Datasets:}
\subsection{Datasets}
We examine 4 binary task datasets and one multi-class task dataset. Following prior research \cite{li-etal-2020-bert-attack}, we randomly sampled up to 1000 examples for each dataset to attack. (QNLI contained 872 examples, thus all were used):

1. AG News - a collection of news articles divided into 4 categories (World, Sports, Business, Sci/Tech). Average text length is 38 words. 

2. SST-2 - Stanford Sentiment Treebank, contains movie reviews labeled for sentiments (positive/negative) by humans. Average text length is 20 words.

3. CoLA - Corpus of Linguistic Acceptability, contains English sentences labeled  grammatical correctness. Average text length is 8 words.

4. QNLI - Stanford Question Answering Dataset, contains question/answer pairs. A classifier must determine whether the context sentence contains the answer to the question. Note that we restrict \textit{VertAttack} to modify the context sentence only. Average text length is 28 words.

5. Rotten Tomatoes (RT) - contains movie reviews from Rotten Tomatoes. Each review is labeled as positive or negative. Average text length is 21 words.

%\noindent \textbf{Classifiers:}
\subsection{Classifiers}
We examine a combination of up to 4 classifiers per dataset. At least 3 classifiers are examined per dataset to measure how well the attack transfers. We look at a combination of transformer models\footnote{We leverage pretrained models via TextAttack: https://github.com/QData/TextAttack} \cite{morris2020textattack}:

1. BERT (base-uncased) - a fine-tuned version of BERT \cite{devlin-etal-2019-bert} on the corresponding dataset. For example, for AG News, the bert-base-uncased model was fine-tuned on the AG News training data.

2. Albert - a fine-tuned version of the ALBERT model \cite{lan2019albert}. ALBERT has a smaller memory footprint than BERT, since it shares weights across layers.

3. RoBERTa - a fine-tuned version of the RoBERTa model \cite{liu2019roberta}. RoBERTa has seen stronger classification results in recent years than BERT, due to choices made during pretraining. 

4. DistilBERT - a fine-tuned version of DistilBERT \cite{sanh2020distilbert}. DistilBERT is a lighter, faster version of BERT which was pretrained using BERT as a teacher for self-supervision. 

%4. Perspective API - Specializing in toxic language classification, this available API assigns various toxic scores to an input text. Following previous research \cite{}, for the offensive dataset, we take a decision of 0.5 toxicity to be offensive. ...

%\noindent \textbf{Metrics:}
\subsection{Metrics}
To calculate the effectiveness of \textit{VertAttack}, we examine 1 quantitative metric and 1 qualitative. 
For quantitative, we measure accuracy:% and accuracy drop:

\begin{equation}
    Accuracy = \frac{\# correctly\_classified}{\# total\_examples}
\end{equation}

\noindent

%\begin{equation}
%    Drop = Original_{Acc.}- Attacked_{Acc.}
%\end{equation}

For qualitative, we measure human ability to understand the text. Specifically, we leverage crowdworkers as judges for the perturbed texts. We ask 3 crowdworkers to label each text (for class) and take the majority vote as a decision.

\section{\textit{VertAttack} Results}

\begin{table}[]
    \centering
    \footnotesize
    \begin{tabular}{c|c||c|c|c|c}
         & & \multicolumn{4}{|c}{Classifiers}\\\hline
         &Feedback & BERT & Albert & Rob. & Disti. \\\hline
         %& BERT & 48.3 & 80.4 & 79.3 & - \\
         %IMDB & Albert & 80.9 & 46.2 & 78.4 & -\\
         %& Rob. & 81.3& 83.0 & 41.2 & -\\\hline
         
         % Yelp
         %& BERT & 44.6 & 77.7 & - & -\\
         %Yelp & Albert & 76.5 & 42.3 & - & -\\\hline
         %& Rob. & & \\
         
         % AG
          \parbox[t]{2mm}{\multirow{4}{*}{\rotatebox[origin=c]{90}{AG}}} & Orig. & 94.2 & 94.2 & 94.7 & - \\
         & BERT & 4.7 & 43.7 & 25.9 & -\\
          & Albert & 60.2 & 8.0 & 31.2 & -\\
         & Rob. & 86.9 & 79.3 & 20.2 & -\\\hline
         
         % SST-2
         \parbox[t]{2mm}{\multirow{4}{*}{\rotatebox[origin=c]{90}{SST-2}}} & Orig. & 92.4 & 92.7 & 94 & - \\
         & BERT & 12.5 & 46.7 & 53.0 & -\\
          & Albert & 53.6 & 13.4 & 57.7 & -  \\
         & Rob. & 50.2 & 51.3 & 13.4 & - \\\hline
         
         % CoLA
         \parbox[t]{2mm}{\multirow{5}{*}{\rotatebox[origin=c]{90}{CoLA}}} & Orig. & 81.2 & 82.9 & 85.7 & 82.5 \\
          & BERT & 5.5 & 29.9 & 35.4 & 33.1 \\
          & Albert & 31.6 & 14.8 & 20.3 & 33.7 \\
         & Rob. & 32.4 & 31.8 & 1.2 & 33.5 \\
         & Disti. & 31.6 & 31.6 & 45.6 & 15.5 \\\hline

         % QNLI
         \parbox[t]{2mm}{\multirow{4}{*}{\rotatebox[origin=c]{90}{QNLI}}} & Orig. & 90.4	& - & 91.7 & 86 \\
          & BERT & 33.5 & - & 67.5 & 60.8 \\
         & Rob. & 62.8 & - & 32.4 & 63.1 \\
         & Disti. & 64.4 & - & 67.8 & 35.6\\\hline
         
         % rotten tomatoes
         \parbox[t]{2mm}{\multirow{4}{*}{\rotatebox[origin=c]{90}{RT}}}& Orig. & 85.4 & 84.8 & 88.6 & - \\
         & BERT & 6.7 & 48.2 & 46.3 & - \\
         & Albert &  46 & 14.7 & 45.2 & -\\
         & Rob. & 56.3 & 40.2 & 25.8 & - \\\hline
    \end{tabular}
    \caption{\textit{VertAttack} results on datasets, accuracy is shown. The second column indicates which classifier was used to give feedback to VertAttack. Orig. = original accuracy without any attack. Rob. = RoBERTa, Disti. = Distilbert. }
    \label{tab:results}
\end{table}

Our main \textit{VertAttack} results are found in Table \ref{tab:results}\footnote{Due to computational intensity of attacks, we opt to test differ combinations of classifiers on the datasets rather than every combination on every dataset.}. The second column indicates which classifier is leveraged for feedback for VertAttack. 
We examine attacks where the feedback and target classifier are the same (diagonal rows), as well as transferability of attacks (non diagonal). Note that the former is the standard measurement in most attack papers. We make the following observations:

% Vert attack strong diagonally (Internal is same as external)
\noindent \textbf{\textit{VertAttack} causes large drops to classifier accuracy.}
Our results demonstrate the effectiveness of \textit{VertAttack} across datasets and classifiers. Specifically, when examining the cases where the feedback classifier is the same as the target classifier, we see up to 90 point drops. In AG News, \textit{VertAttack} is able to drop BERT from 94.2\% to 4.7\%, Albert from 94.2 to 8.0, and RoBERTa from 94.7 to 20.2, which averages to 83 points. Similar drops from \textit{VertAttack} are seen in the other datasets as well: SST-2 averages 80 points, CoLA averages 74 points, QNLI averages 56 points, and Rotten Tomatoes averages 71 points. Overall, these results support VertAttack's strength in fooling classification systems.

% transferability causes close to 50% accuracy drops 
\noindent \textbf{VertAttack's attacks transfer to other classifiers. } 
Though not as strong, we find \textit{VertAttack} to be successful even in cases of transferability. In the most effective case (the CoLA datasets), the transfer attacks cause an average drop of 51 points (max: 65, min: 40.1). These drops are detrimental to text classifiers' effectiveness and reliability. Slightly lesser drops are seen for SST-2, AG News, and Rotten Tomatoes which causes drops around 40 points on average. Finally, classifiers on the QNLI dataset see drops of 25 when the feedback classifier differs. In even the final cases, the attacks is a hinderance to classification methods and highlight their inability to process text as effectively as humans.

% QNLI models hardest to attack
\noindent \textbf{QNLI models most resilient to attack.}
Unlike the other datasets, which saw at least 1 classifier drop below 20\% classification accuracy, QNLI classifiers dropped to only 32\% in the lowest. This might be due to the difficulty of attacking multi-text inputs. We limited \textit{VertAttack} to only attack the hypothesis and not the premise. We would most likely see a drop in accuracy if premise is allowed to be attacked as well, but we restricted to the hypothesis for a more realistic model where a user is proposing a hypothesis to a model's premise.  

% Which internal is best?
%...

\begin{table}[]
    \footnotesize
    \centering
    \begin{tabular}{c|c|c|c|c}

          %& \multicolumn{4}{|c}{Classifiers}\\\hline
         & BERT & Albert & Rob. & Disti. \\\hline
         %& BERT & 48.3 & 80.4 & 79.3 & - \\
         %IMDB & Albert & 80.9 & 46.2 & 78.4 & -\\
         %& Rob. & 81.3& 83.0 & 41.2 & -\\\hline
         
         % Yelp
         %& BERT & 44.6 & 77.7 & - & -\\
         %Yelp & Albert & 76.5 & 42.3 & - & -\\\hline
         %& Rob. & & \\
         
         %average drops
         
         Same & 76.1 & 75.9 & 72.3 & \textbf{58.7}\\
         Diff. & \textbf{35.7} & 43.3 & 45.4 & 39.06\\
         All & 48.3 & 53.3 & 53.8 & \textbf{44.7} \\\hline
         \hline
    \end{tabular}
    \caption{Average drops of \textit{VertAttack} against the corresponding classifier across all datasets.  Three averages are shown:  ``Same'' indicates the average of the attacks where the feedback classifier was the same as the attacked. ``Diff.'' indicate the set of attacks where the feedback classifier differed from the attacked.``All'' is the average for all drops against the classifier. Bold values indicate lowest drops.}
    \label{tab:averagedrops}
\end{table}

% RoBERTa most robust except for a few distilbert cases
\noindent \textbf{BERT and DistilBert show strength as most robust classifiers examined.}
To investigate resilience against VertAttack, we calculate three averages for each classifier, seen in Table \ref{tab:averagedrops}: 1. The classifier used by \textit{VertAttack} for feedback is the same as the target classifier (Same), 2. The classifier used by \textit{VertAttack} is \textbf{different} than the target classifier (Diff.), 3. Inclusion of both 1 and 2 (All). 
Each score corresponds to the drop in accuracy against VertAttack. Thus, for resiliency, classifiers would like to have a lower drop in accuracy. We can see that DistilBert has the lowest drops in two cases (Same, All), while BERT has the lowest for the third (Diff.). However, BERT is examined in all 5 datasets, while DistilBert is only examined in 2. Thus, no final decision can be noted on most resilient between the two.

\section{Human Study}\label{sect:humanstudy}
To investigate humans' understanding of VertAttack's texts, we employed human crowdworkers to label a sampled set of texts which were perturbed by VertAttack. Specifically, we randomly sampled 100 of the 1000 texts from the Rotten Tomatoes dataset. We then asked crowdworkers to read the text and decide the sentiment of the text (positive or negative). For each text, we employed 3 crowdworkers\footnote{Amazon Mechanical Turk}, and took the majority vote of the labels. It should be noted that no instructions to read the texts vertically were given. More information on the instructions can be found in Appendix \ref{sect:hstudy}. 

\begin{table}[]
    \footnotesize
    \begin{subtable}[h]{0.45\columnwidth}
    \centering
    \begin{tabular}{cc|c|c}
        \multicolumn{4}{c}{\textit{\textbf{VertAttack}}}\\
         &  & \multicolumn{2}{c}{Actual} \\
         & & + & - \\\hline
         \parbox[t]{2mm}{\multirow{2}{*}{\rotatebox[origin=c]{90}{Pred.}}} & + & 41 & 16\\ 
         & - & 7 & 36 \\\hline
    \end{tabular}
    \end{subtable}
    \hfill
    \begin{subtable}{0.45\columnwidth}
        \centering
    \begin{tabular}{cc|c|c}
         \multicolumn{4}{c}{\textbf{Original}}\\
         &  & \multicolumn{2}{c}{Actual} \\
         & & + & - \\\hline
         \parbox[t]{2mm}{\multirow{2}{*}{\rotatebox[origin=c]{90}{Pred.}}} & + & 40 & 11\\ 
         & - & 8 & 41 \\\hline
    \end{tabular}
    \end{subtable}
    \caption{Confusion Matrices of human study results. Participants labeled 100 perturbed RT texts as positive (+) or negative (-) sentiment. Each text received 3 votes, a majority vote was taken. }
    \label{tab:humanstudymatrix}
\end{table}

The confusion matrix of results is in Table \ref{tab:humanstudymatrix}. Humans were able to identify sentiment correctly, 77\% of the time, far greater than the 7 - 26\% of the automated classifiers. This confirms that unlike the automated classifiers, humans are well prepared to read text in non-traditional manners. 

For comparison, we also ran the same study with on the original, unperturbed 100 texts.  This is also in Table \ref{tab:humanstudymatrix} under the ``Original'' subtable. Humans are able to do slightly better on the unperturbed texts achieving an accuracy of 81\%. However, VertAttack's percentage is only 4 points below (77\%). This highlights that human misclassifications on VertAttack's texts have more to do with the difficulty of some of the texts rather than due to perturbation.

\section{Comparisons with other attacks}
To further investigate how \textit{VertAttack} performs in the adversarial text space, we compare to two other attacks, BERT-ATTACK \cite{li-etal-2020-bert-attack} and Textbugger \cite{Li2019Textbugger}\footnote{TextAttack was leveraged to simulate these attacks: github.com/QData/TextAttack}. BERT-Attack is similar to \textit{VertAttack} as it is a word based attack. To select a word, BERT-ATTACK finds the importance score of a word by masking each word (one at a time) and comparing to the original logits. 
For replacement, BERT-ATTACK relies on BERT to give suggestions via its MLM training.  Textbugger is a character based attack which tests inserting, deleting, swapping, or substituting characters. We run both attacks on the same 1000 examples from the Rotten Tomatoes dataset. The results can be seen in Table \ref{tab:comparisons}.

Overall, we find that BERT-ATTACK causes greater drops when the feedback classifier is the same as the attacked classifier, but \textit{VertAttack} transfers better. Textbugger is weaker in both cases. Specifically,  when the feedback classifier is the same (diagonal values), BERT-ATTACK causes classifiers to average 9.5\% accuracy compared to VertAttack's 15.7\% and Textbugger's 33.5\%. However, for transferability (non diagonal values), \textit{VertAttack} causes classifiers to average 47\% accuracy, 19 points less than BERT-ATTACK's average of 66.5\% and 31 points less than Textbugger's average of 78.1. Furthermore, when taking the overall averages (all cells) \textit{VertAttack} drops classifiers to 36.6\% accuracy while BERT-ATTACK averages 47.5\% and Textbugger averages 63.2\%. 
%Thus we can confirm further effectiveness of VertAttack. 

\begin{table}[]
    \footnotesize
    \centering
    \begin{tabular}{c|c|c|c|c}
          & & \multicolumn{3}{|c}{Classifiers}\\\hline
          & & BERT & Albert & RoBERTa \\\hline
           & Original & 85.4 & 84.8 & 88.6 \\\hline 
         % rotten tomatoes
         % vertattack
          \parbox[t]{2mm}{\multirow{3}{*}{\rotatebox[origin=c]{90}{Vert A.}}}  & BERT & \textbf{6.7} & 48.2 & 46.3  \\
          & Albert &  \textit{46} & 14.7 & \textit{45.2} \\
          & RoBERTa & 56.3 & \textit{40.2} & 25.8  \\\hline

         % BERT-ATTACK
          \parbox[t]{2mm}{\multirow{3}{*}{\rotatebox[origin=c]{90}{Bert A.}}} & BERT &22.9 & 52.3 & 74.8 \\
          & Albert & 79 & \textbf{1.9} & 78.7\\
          & RoBERTa & 66.6 & 47.3 & \textbf{3.6} \\\hline

        % textbugger
        \parbox[t]{2mm}{\multirow{3}{*}{\rotatebox[origin=c]{90}{Textb.}}} & BERT 
          & 46.2 & 52.3 & 74.8 \\
          & Albert & 85.8 & 16.1 & 91.6\\
          & RoBERTa & 74.1 & 56.9 & 38.2 \\\hline
          
    \end{tabular}
    \caption{\textit{VertAttack} compared with BERT-Atttack and Textbugger. The second column indicates which classifier was used to give feedback to the attacks. Bold values indicate stronger attacks against that classifer. Italic values indicate strongest transfer attack.}
    \label{tab:comparisons}
\end{table}

\section{Malicious Use - Offensive Language}

\begin{table}[]
    \centering
    \footnotesize
    \begin{tabular}{c|c||c|c|c}
         & & \multicolumn{3}{|c}{Classifiers}\\\hline
         &Feedback & BERT & Albert & XLNet \\\hline
         
         % OLID
         %\parbox[t]{2mm}{\multirow{3}{*}{\rotatebox[origin=c]{90}{OLID }}}
         & Original & 76.7 & 78.3 &  78.3 \\
         & BERT & 1.3 & 23.8 & 27.5 \\
         & Albert & 20 & 0 & 26.7\\
         & XLNet & 12.9 & 17.1 & 0.8\\\hline
    \end{tabular}
    \caption{\textit{VertAttack} results on OLID dataset, on the OFF labeled (Offensive Language). Accuracy is shown. The second column indicates which classifier was used to give feedback to \textit{VertAttack}. }
    \label{tab:OFFresults}
\end{table}

To confirm the main results and demonstrate how \textit{VertAttack} may be used maliciously, we apply \textit{VertAttack} to ``offensive'' texts. We take a subset of OLID's \cite{zampierietal2019OLID} test set, labeled OFF (offensive). This results in 260 texts. We leveraged pretrained classifiers from Huggingface \footnote{https://huggingface.co/mohsenfayyaz}, trained on OLID training data. We examine 3 variations of transformer models, BERT, Albert, and XLNet \cite{yang2019xlnet}. The full results are in Table \ref{tab:OFFresults}. 

\textit{VertAttack} is able to greatly reduce the classification accuracy for all three models. When the feedback classifier is the same as the target, the accuracy drops to ~1\% or lower. When the classifiers differ, the accuracy is also low, in the range 13 - 28\%. These results demonstrate how the attack can cause issues on popular social media websites which leverage automated classifiers to help curb offensive language. 
\section{Effect on OCR + Classifier}
To guarantee the preservation of whitespace, we can write text to an image (as done in the human study). The question arises of how a classifier which leverages OCR to extract text from images would fare. We test this by first converting the modified text into an image using the PIL library\footnote{https://pypi.org/project/Pillow/}. Next, we use Tesseract OCR\footnote{https://github.com/tesseract-ocr/tesseract} to extract the text from the image and classify it. We test this on Rotten Tomatoes. The feedback and target classifiers use the text segmenter (Section \ref{sect:defense}). The results can be found in Table \ref{tab:ocrresults}. We include a simple majority class baseline for comparison.

\begin{table}[]
    \centering
    \footnotesize
    \begin{tabular}{c|c||c|c|c}
         & & \multicolumn{3}{|c}{Classifiers}\\\hline
         &Feedback & BERT & Albert & RoBERTa \\\hline

         % rotten tomatoes
         & Original & 85.4 & 84.8 & 88.6 \\\hline

          \parbox[t]{2mm}{\multirow{3}{*}{\rotatebox[origin=c]{90}{None}}}& BERT & 6.7 & 48.7 & 50  \\
         & Albert &  47.7 & 13.6 & 48.7 \\
         & RoBERTa & 44.8 & 45.5 & 9.4 \\\hline

         \parbox[t]{2mm}{\multirow{3}{*}{\rotatebox[origin=c]{90}{OCR}}}& BERT & 40.5 & 47.3 & 48.2  \\
         & Albert & 48.4 & 35.7 & 49.2 \\
         & RoBERTa & 45.6 & 44.1 & 37.7 \\\hline

          & Maj. Class & \multicolumn{3}{|c}{53.3} \\
         %& Neg. & \multicolumn{3}{|c}{46.7}\\\hline

    \end{tabular}
    \caption{Accuracy results on RT dataset when images containing VertAttack modified text are converted to text (via OCR) and classified. ``None'' refers to the original accuracy with no conversion to image and back via OCR. 
    Second column indicates which classifier was used for attack feedback. ``Maj. Class'' indicates a simple baseline which always predicts the majority class. }
    \label{tab:ocrresults}
\end{table}

For OCR, we see accuracy increase in the cases when the target and feedback classifier are the same. For example, Albert classification changes from 13.6 to 35.7. 
When feedback and target classifiers differ, the accuracy is similar to the original attacked accuracy. All accuracies are below the simple majority class baseline of 53.3.  Thus, even though OCR increase accuracy, it is still detrimental for a classifier. Furthermore, VertAttack could be further modified to target a classifier which includes OCR in the pipeline.

\section{Initial Defenses}\label{sect:defense}

We investigate some initial steps automated classifiers might take to mitigate VertAttack's effectiveness. 
Since \textit{VertAttack} introduces  whitespace, simple solutions might be to reduce that whitespace. Thus, we look at three different approaches. First, we simply remove extraneous whitespace and limit at most 1 space between each token, denoted as \textbf{Simple}. Second, we leverage a text segmentation library\footnote{grantjenks.com/docs/wordsegment/} to remove whitespace and re-combine words, denoted as \textbf{Segment}. Finally, we assume the classifier has learned the algorithm for VertAttack and thus reverses it. That is, the classifier attempts to recombine vertical characters into words before classification. This is denoted as \textbf{Reverse}. The full algorithm can be found in the appendix (Appendix \ref{sect:reversealg}).

\subsection{Simple + Segment}
For the first two approaches, we run them on the original attacked Rotten Tomato (RT) texts (from Table \ref{tab:results}). 
We then modify \textit{VertAttack} to have this information during its attacks as feedback, as changing the preprocessing method during classification puts the attack at a natural disadvantage since the feedback is no longer as reliable.
The full results of these experiments are in Table \ref{tab:preprocessresults}. 
%
%Note that the title rows (``\textit{VertAttack} - X'') indicate which preprocessing approach was leveraged by the feedback classifier and the left most column indicates which approach was used by the attacked classifiers. For example, in the final rows of results (``\textit{VertAttack} - Segmenter'') VertAttack's feedback classifiers use the text segmenter in the model. 

\begin{table}[]
    \centering
    \footnotesize
    \begin{tabular}{c|c||c|c|c}
         & & \multicolumn{3}{|c}{Classifiers}\\\hline
         &Feedback & BERT & Albert & RoBERTa \\\hline
         %& BERT & 48.3 & 80.4 & 79.3 & - \\
         %IMDB & Albert & 80.9 & 46.2 & 78.4 & -\\
         %& RoBERTa & 81.3& 83.0 & 41.2 & -\\\hline

         % rotten tomatoes
         & Original & 85.4 & 84.8 & 88.6 \\\hline

          & \multicolumn{4}{|c}{\textit{VertAttack} - None}\\\hline
         %No Preprocessing
%         \parbox[t]{2mm}{\multirow{3}{*}{\rotatebox[origin=c]{90}{None}}}& BERT & 6.7 & 48.2 & 46.3  \\
%         & Albert &  46.0 & 14.7 & 45.2 \\
%         & RoBERTa & 56.3 & 40.2 & 25.8 \\\hline

          %simple vs none
         \parbox[t]{2mm}{\multirow{3}{*}{\rotatebox[origin=c]{90}{Simple}}}& BERT & 6.7 & 48.7 & 50.0  \\
         & Albert &  46.0 & 29.7 & 47.6 \\
         & RoBERTa & 56.3 & 38.1 & 59.8 \\\hline

         %segment vs none
         \parbox[t]{2mm}{\multirow{3}{*}{\rotatebox[origin=c]{90}{Seg.}}}& BERT & 37.8 & 49.6 & 53.8  \\
         & Albert & 45.4 & 49.2 & 51.1 \\
         & RoBERTa & 62.3 & 43.8 & 62.8 \\\hline

        & \multicolumn{4}{|c}{\textit{VertAttack} - Simple}\\\hline
        %simple vs simple
         \parbox[t]{2mm}{\multirow{3}{*}{\rotatebox[origin=c]{90}{Simple}}} & BERT & 6.7 & 48.7 & 50  \\
         & Albert &  47.7 & 13.6 & 48.7 \\
         & RoBERTa & 44.8 & 45.5 & 9.4 \\\hline

        & \multicolumn{4}{|c}{\textit{VertAttack} - Segmenter}\\\hline
         %segment vs segment
         \parbox[t]{2mm}{\multirow{3}{*}{\rotatebox[origin=c]{90}{Seg.}}}& BERT & 10.0 & 44.7 & 53.6  \\
         & Albert & 49.3 & 4.7 & 53.6 \\
         & RoBERTa & 41.7 & 41.8 & 8.2 \\\hline
         
    \end{tabular}
    \caption{\textit{VertAttack} results on RT dataset with different whitespace preprocessing present, accuracy is shown.
    First column indicates which method the classifier used: Simple - remove all extraneous spaces in input text, Seg. - leverage word segmenter to process the input text. 
    Second column indicates which classifier was used to give feedback to VertAttack.
    ``\textit{VertAttack} - X'' indicates which method \textit{VertAttack} used with classifier feedback. }
    \label{tab:preprocessresults}
\end{table}

We observe that when \textit{VertAttack} includes a preprocessing method for feedback that is different than what the attacked classifier uses (``\textit{VertAttack} - None''), the attack suffers. For example, examining the diagonal results, the simple preprocessing is able to raise Albert's classification accuracy from 14.7 to 29.7. The word segmentation approach raises it even higher (to 49.2). Similar results are seen across the table. The transferability results (feedback classifier differs from final classifier) also generally increase, but not nearly as strong. This follows as \textit{VertAttack} is modifying texts based on a classifier that differs in preprocessing and hence the attack becomes a transferability problem itself. 

When \textit{VertAttack} has the same method in its feedback classifier, then the approaches are not as fruitful (``\textit{VertAttack} - Simple'', ``\textit{VertAttack} - Segmenter''). Again with Albert (on the diagonal), we actually see a decrease in classification accuracy from 14.7 to 13.6 for \textbf{Simple} and down to 4.7 for \textbf{Segmentation}. This indicates the importance of the feedback classifier as it can strongly affect VertAttack's perception of a strong attack and the importance of whitepace preprocessing for classifiers if the attacker is not prepared.

\subsection{Reverse}\label{sect:reverse}
\begin{table}[]
    \centering
    \footnotesize
    \begin{tabular}{c|c||c|c|c}
         & & \multicolumn{3}{|c}{Classifiers}\\\hline
         &Feedback & BERT & Albert & RoBERTa \\\hline
         %& BERT & 48.3 & 80.4 & 79.3 & - \\
         %IMDB & Albert & 80.9 & 46.2 & 78.4 & -\\
         %& RoBERTa & 81.3& 83.0 & 41.2 & -\\\hline

         % rotten tomatoes
         & Original & 85.4 & 84.8 & 88.6 \\\hline

         %No Preprocessing
         \parbox[t]{2mm}{\multirow{3}{*}{\rotatebox[origin=c]{90}{Reverse}}}& BERT & 84.4 & 84.2 & 88.4  \\
         & Albert &  82.6 & 84.3 & 87.8 \\
         & RoBERTa & 86 & 82.6 & 87.3 \\\hline

    \end{tabular}
    \caption{\textit{VertAttack} results on RT dataset when the classifier reverse-engineers VertAttack, accuracy is shown.}
    \label{tab:reverseresults}
\end{table}

The \textbf{Reverse} preprocessing results can be found in Table \ref{tab:reverseresults}. As can be observed, the algorithm is able to strongly combat VertAttack, increasing the accuracy from 6 - 24 to 84 - 87. However, we observe that it is not able to mitigate it entirely, as some texts are entirely written vertically and the algorithm is not able to distinguish when new lines of words begin. We next introduce an augmentation to VertAttack to combat the \textbf{Reverse} algorithm.

\section{Enhancing VertAttack with Chaff}\label{sect:chaff}
As demonstrated, if the classifier knows this type of attack is occurring, it can strongly mitigate it by reversing the algorithm. Thus, we enhance VertAttack by introducing chaff. Specifically, rather than inserting only whitespace vertically, an alphabet character has a chance of being inserted. This occurs at a probability \textit{p}. For example, if $p = 10$, then there is a 10\% probability that rather than whitespace, a character is inserted in the vertical lines. Note that to preserve readability we do not allow this for whitespace next to perturbed words (nor original whitespace). 
\begin{table}[]
    \centering
    \footnotesize
    \begin{tabular}{c|c||c|c|c}
         & & \multicolumn{3}{|c}{Classifiers}\\\hline
         &Feedback & BERT & Albert & RoBERTa \\\hline
         %& BERT & 48.3 & 80.4 & 79.3 & - \\
         %IMDB & Albert & 80.9 & 46.2 & 78.4 & -\\
         %& RoBERTa & 81.3& 83.0 & 41.2 & -\\\hline

         % rotten tomatoes
         & Original & 85.4 & 84.8 & 88.6 \\\hline

         %No chaff
          & \multicolumn{4}{|c}{$p$ = 0\%}\\\hline
          \parbox[t]{2mm}{\multirow{3}{*}{\rotatebox[origin=c]{90}{None}}}& BERT & 6.7 & 48.2 & 46.3  \\
         & Albert &  46.0 & 14.7 & 45.2 \\
         & RoBERTa & 56.3 & 40.2 & 25.8 \\\hline
         \parbox[t]{2mm}{\multirow{3}{*}{\rotatebox[origin=c]{90}{Reverse}}}& BERT & 84.4 & 84.2 & 88.4  \\
         & Albert &  82.6 & 84.3 & 87.8 \\
         & RoBERTa & 86 & 82.6 & 87.3 \\\hline

        % p = 5%
        & \multicolumn{4}{|c}{$p$ = 5\%}\\\hline
          \parbox[t]{2mm}{\multirow{3}{*}{\rotatebox[origin=c]{90}{None}}}& BERT & 6.4 & 48.3 & 46.1  \\
         & Albert & 46.8 & 15.9 & 44.9 \\
         & RoBERTa & 57.7 & 41.3 & 24.6 \\\hline
         \parbox[t]{2mm}{\multirow{3}{*}{\rotatebox[origin=c]{90}{Reverse}}}& BERT & 76.4 & 78.1 & 81.1  \\
         & Albert & 75.8 & 75.7 & 82.0 \\
         & RoBERTa & 77.9 & 76.3 & 78.6\\\hline
         
         % p = 10%
        & \multicolumn{4}{|c}{$p$ = 10\%}\\\hline
          \parbox[t]{2mm}{\multirow{3}{*}{\rotatebox[origin=c]{90}{None}}}& BERT & 6.0 & 49.1 & 46.3   \\
         & Albert & 46.3 & 17.0 & 44.4 \\
         & RoBERTa & 57.33 & 42.0 & 24.4  \\\hline
         \parbox[t]{2mm}{\multirow{3}{*}{\rotatebox[origin=c]{90}{Reverse}}}& BERT & 64.8 & 70.7 & 71.6   \\
         & Albert & 68.2 & 64.7 & 76.2 \\
         & RoBERTa & 73.7 & 71.5 & 67.4  \\\hline

        % p = 20%
        & \multicolumn{4}{|c}{$p$ = 20\%}\\\hline
          \parbox[t]{2mm}{\multirow{3}{*}{\rotatebox[origin=c]{90}{None}}}& BERT & 5.9 &  48.4 & 46.6  \\
         & Albert & 45.3 & 18 & 45.2 \\
         & RoBERTa & 57.7 & 42.2 & 24.2 \\\hline
         \parbox[t]{2mm}{\multirow{3}{*}{\rotatebox[origin=c]{90}{Reverse}}}& BERT &  48.7 & 63.2 & 62.4  \\
         & Albert & 60.8 & 47.1 & 67.9  \\
         & RoBERTa & 67.1 & 69.7 & 50.2 \\\hline

         % p = 30%
        & \multicolumn{4}{|c}{$p$ = 30\%}\\\hline
          \parbox[t]{2mm}{\multirow{3}{*}{\rotatebox[origin=c]{90}{None}}}& BERT & 5.8 & 49.2 & 47.6  \\
         & Albert & 44.7 & 19.6 & 44.3 \\
         & RoBERTa & 55.5 & 42.3 & 23.7\\\hline
         \parbox[t]{2mm}{\multirow{3}{*}{\rotatebox[origin=c]{90}{Reverse}}}& BERT & 39.8 & 59.3 & 58.2  \\
         & Albert & 58.1 & 40.1 & 64.5 \\
         & RoBERTa & 63.8 & 65.8 & 40.5\\\hline

         % p = 60%
        & \multicolumn{4}{|c}{$p$ = 60\%}\\\hline
          \parbox[t]{2mm}{\multirow{3}{*}{\rotatebox[origin=c]{90}{None}}}& BERT & 6.2 & 48.5 & 47.4  \\
         & Albert & 45.2 & 21.0 & 43.7 \\
         & RoBERTa & 55.5 & 42.3 & 23.7 \\\hline
         \parbox[t]{2mm}{\multirow{3}{*}{\rotatebox[origin=c]{90}{Reverse}}}& BERT & 27.7 & 60.1 & 55.0  \\
         & Albert & 57.5 & 28.9 & 63.2 \\
         & RoBERTa & 59.7 & 64.1 & 35.9 \\\hline
         
    \end{tabular}
    \caption{\textit{VertAttack} results on RT dataset when chaff is added in (described in Section \ref{sect:chaff}). ``None'' means no preprocessing is used and ``Reverse'' is the classifier attempting to reverse engineer VertAttack.}
    \label{tab:fullpchaffresults}
\end{table}

\begin{table}[]
    \centering
    \begin{tabular}{c|c|c|c}
    & \multicolumn{3}{|c}{\# of correct responses}\\\hline
         & >= 1 & >= 2 & =3 \\\hline
         Original & 94 & 81 & 49 \\
         VertAttack & 92 & 77 & 47 \\
         Chaff $p = 30$ & 83 & 47 & 23 \\\hline
    \end{tabular}
    \caption{Human results for all three text variations. The values indicate the percentages of texts correctly classified by at least X humans where X is indicated in the column header. Original and VertAttack are the same values from Table \ref{tab:humanstudymatrix}. Chaff $p = 30$ indicates that chaff is added to the perturbed text at 30\% rate.  }
    \label{tab:pchaffhumanresults}
\end{table}

 %The entire results can be found in Appendix \ref{sect:fullpchaffresults}, which also includes the chaff against no prepocessing as well. 

We test chaff for $p = \{5, 10, 20, 30, 60\}$. The main results  against the \textbf{Reverse} algorithm (Section \ref{sect:reverse}) are in Table \ref{tab:fullpchaffresults}.
We find that this enhancement hinders the ability to reverse the attack. This is because \textbf{Reverse} is not able to identify non-perturbed characters. For example, when $p=30$ BERT's accuracy drops from 85 to 40. This is 44 points lower than when the \textbf{Reverse} is applied to $p=0$ (no chaff). Similar trends are seen for Albert and RoBERTa as well. As $p$ increases, we find greater accuracy drops. This points to the reverse algorithm becoming less able to avoid the random inserted text.

We verify that readability is maintained, following the same process in the main human study (Section \ref{sect:humanstudy}). Table \ref{tab:pchaffhumanresults} compares human evaluations of adding in chaff at a rate of 30\%. We see a drop in correct responses but at least 1 human is able to correctly identify the sentiment in at least 83\% of the texts. 

% The results can be seen in the Appendix (Table \ref{tab:pchaffhumanresults}). We observe some slight drop in human classification when chaff is increased, but text is understood by at least 1 human in 83\% of texts. 

This enhancement further demonstrates \textit{VertAttack} as a strong representation of how humans can adjust to combat automatic classifiers.

\section{Related Work}\label{sect:relatedwork}
Here we examine some of the other current SOTA attacks. We examine both word-based attacks and character-based attacks as \textit{VertAttack} shares some characteristics with both. 

% sota attacks
% word based
\noindent \textbf{Word-Based Attacks:}
Like \textit{VertAttack}, current black-box SOTA word-based attacks attack a classifier by receiving feedback from that classifier. This feedback is in the form of label probabilities \cite{hsieh-etal-2019-robustness}, or the logits of the classifier \cite{Li2021SearchingFA}.
Black-box, word-based attacks follow similar steps to \textit{VertAttack}. First, they choose tokens for replacement, and then they leverage a tool to choose a replacement. This could be a transformer model \cite{li-etal-2020-bert-attack}, a lexicon like WordNet \cite{ren-etal-2019-generating}, or word embeddings \cite{hsieh-etal-2019-robustness}. 
Unlike, \textit{VertAttack} current word-based attacks only operate in the horizontal space. That is, all words chosen for replacement are substituted for that word in place. Their goal is to find words which a classifier does not know well enough to make a correct classification. 
Thus, \textit{VertAttack} is set apart by operating in the vertical space. Furthermore, \textit{VertAttack} does not replace the selected word, thus meaning is more easily preserved. 

% character based
\noindent \textbf{Character-Based Attacks:}
Another common type of SOTA attack are character-based attacks which change text at the character level. 
These attacks generally aim to be more transferable than word attack and thus do not receive feedback from a classifier. Instead, the changes are applied at a random chance throughout the text. 
For example, whitespace might be removed \cite{Grndahl2018AllYN} or added or standard, English characters might be replaced with non-standard similar looking characters (e.g ``a'' $\rightarrow$ ``@'')\cite{eger-etal-2019-text}. Both cases try to cause classifiers to see words as out-of-vocabulary. 
One downside is that character-level attacks can be mitigated more easily with proper preprocessing \cite{rusert2022robustness}. 
\textit{VertAttack} is similar, in that it focuses on the characters of a word, however, \textit{VertAttack} uses an internal classifier for feedback. Furthermore, due to the positioning of the characters, \textit{VertAttack}'s changes are harder to correct with preprocessing of text. 

Whitespace attacks have also been shown to be effective against LLMs. \citet{cai2023evade} find that adding a whitespace before a comma in a text can fool a classifier to misclassify a text as human-generated instead of machine-generated. This attack, \textit{SpaceInfi}, differs from \textit{VertAttack} since it only focuses on this specific classification task. Furthermore, it adds a single space next to commas. In our experimentations, we focus on classification tasks where syntactic structure is less important.  
%\noindent \textbf{Vertical Texts:} As noted, no prior attacks nor papers on text classifier robustness have studied vertical texts. However, one related area has seen similar research. Specifically, 

% classification of vertical text? (if such research exists)

\section{Conclusion}
We presented a new attack which exploits current classifiers' inability to understand text written vertically. Mimicking a human, \textit{VertAttack} perturbs text by rewriting words in a vertical manner which humans are able to understand, but classifiers are not. We find drops in classification up to 86 points. %\textit{VertAttack} also transfers well seeing average accuracy drops of 51. 

Furthermore, \textit{VertAttack} produces texts which humans can understand. Human crowd workers verified this by labeling 77\% perturbed texts correctly, compared to 81\% of non perturbed texts. %This points to misclassification caused by difficulty of texts and not perturbations. 

When compared to other attacks, \textit{VertAttack} causes stronger drops when transferability of attacks is included. \textit{VertAttack} drops classifiers to 36.6\% accuracy compared to 46.5\% of BERT-Attack and 63.2\% of Textbugger.

We explored initial results on how \textit{VertAttack} affects classifiers with OCR. We found that these classifiers are more robust, but still vulnerable.

Finally,  We investigated initial defenses against \textit{VertAttack} and found that the methods are able to mitigate the attack as long as \textit{VertAttack} does not enhance with chaff. 

\vspace{0.3cm}

\noindent Every experiment shows \textit{VertAttack}'s ability to maintain 
readability and cause large accuracy drops in multiple classifiers. We also find humans do know the meaning in the attacked text. Hence, the overall results will be useful for future research.
%\textit{VertAttack} causes strong drops and maintains meaning. Our hope is \textit{VertAttack} may inspire future research into addressing this limitation of classifiers be robust against attacks inspired by humans.

\section{Limitations}
Here we note some limitations with our method and with our experiments. These limitations should be kept in mind when working and expanding on \textit{VertAttack} so that they addressed or noted:

\noindent \textbf{Websites are not guaranteed to preserve formatting of text produced by VertAttack.}
\textit{VertAttack} produces text in which targeted words are vertically perturbed. It does this by adding in multiple newlines characters and padded whitespace to preserve readability. However, not all websites are guaranteed to preserve this additional whitespace. Some may completely remove extra newlines which will cause the produced text to greatly drop in readability. One solution to this is leveraging a module to write the text into an image (as seen in the examples (Figure \ref{fig:examples}). With an image, the formatting of text will be honored and readable to humans. Furthermore, this adds another layer to the attack as text would first need to be processed from the image for classification. However, not all websites allow images, and thus it is a noted limitation to be remedied in the future. 

\noindent \textbf{Our attacks focused exclusively on transformer classification models.} Though transformers are the current kings of classification, not all websites might have the resources to employ these types of models and thus investigation into simpler models may be useful to confirm VertAttack's effectiveness. However, generally non-transformer models have struggled against adversarial attacks and in the past, and there seems to be no reason why they would fare any better against VertAttack. 

\noindent \textbf{Greedy word selection is time consuming.} The selection method is the least efficient part of VertAttack. As noted, many previous attacks have leveraged a similar method (Section \ref{sect:methods}). This is due to lack of classifier knowledge in blackbox approaches, thus most tokens need to be checked in selection. However, there do exist more efficient approaches. For example, some style transfer algorithms use attention mechanisms to find the most important words \cite{ijcai2019p732}. Thus, \textit{VertAttack} could be further improved by improving the selection algorithm.

%\noindent \textbf{Experiments are exclusively in English datasets.} Though other languages share similar properties to English, there is a lack in experiments in other languages. Future work would verify the potency of \textit{VertAttack} in other languages, so classifiers in other languages could examine their own robustness. 
\section{Ethical Considerations}
By simulating adversarial attacks, such as \textit{VertAttack}, concerns can arise over ethical implications. For example, introducing such a method might allow malicious users to more easily introduce harmful texts into websites and other spaces. This is a further concern as, for research, we make code and algorithms publicly available. This needs to be considered when introducing and studying any adversarial attack. However, we believe that in spite of the above possible wrongful uses, \textit{VertAttack} can be helpful in studying both robustness and future understanding tasks of text classification systems. This is further emphasized as humans can naturally perform this attack and there is no dataset which collects these attacks done by humans. Hence, \textit{VertAttack} provides a way to simulate and further study such attacks. Through this simulation, classifiers, defenses, and other related NLP systems can benefit in a public space. Our hope is not that this algorithm is ever used for malicious purposes, but to improve the aforementioned systems. Thus, we believe the benefits to outweigh any risks.

\bibliography{main}
\bibliographystyle{acl_natbib}

\newpage
\newpage
\appendix
\onecolumn
\begin{figure*}
    \centering
    \includegraphics[width=\textwidth]{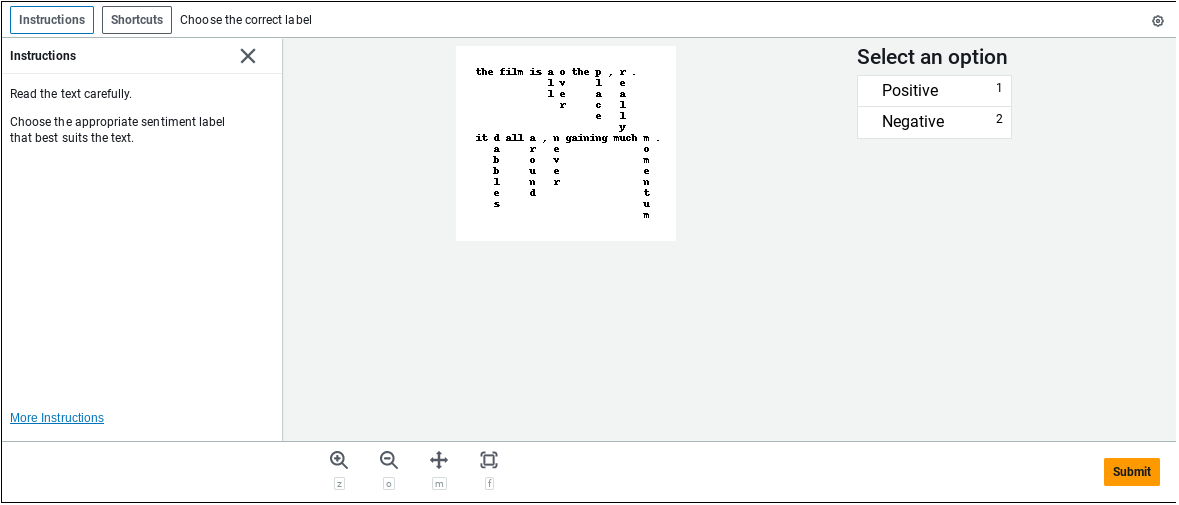}
    \caption{Instructions shown to Amazon Mechanical Turk crowdworkers.}
    \label{fig:instructions}
\end{figure*}

\newpage
\twocolumn
\section{Human Study Details}\label{sect:hstudy}

For the human study we leveraged Amazon Mechanical Turk crowdworkers to annotate sentiment on Rotten Tomatoes text which were perturbed by VertAttack. The instructions provided to the participants can be seen in Figure \ref{fig:instructions}. As can be seen, no instructions to read the text vertically were given. For each annotation of text, crowdworkers were paid \$0.08. Each text received 3 annotations. As AMT does presents each text as a separate task, the 3 annotators for 1 text were rarely the same annotators for another task, thus annotator agreement was not calculated. 

To present the texts, we leverage the PIL library in python to write the texts into simple images. An example of this can be seen in the example images (Figure \ref{fig:examples}). We chose to push the text onto images to avoid any website dependent presentation of the text (e.g. the worker viewer the text on a desktop versus on a phone). 

\section{Reverse Algorithm}\label{sect:reversealg}
\begin{algorithm}
\caption{\textbf{Reverse}}\label{alg:reverse}
\begin{algorithmic}
\Require $\text{Perturbed Text}$
\Ensure $\text{Preprocessed Text}$
\State $\text{Split\_Text} \gets \text{Text.Split(`$\setminus$n')}$
\State $Drop_{Max} \gets 0, i \gets 0, j \gets 0$
\State $Top\_Line \gets 0$
\While{$i \leq \text{Split\_Text.length()}$}
    \State $cur\_line = Split\_Text[i]$
    \If{$length(word) \in cur\_line > 1 $}
        \State $\text{update previous top line, add to final text}$
        \State $Top\_Line \gets i$
    \Else 
        \State $\text{store characters at positions}$
    \EndIf
    \State $i \gets i + 1$
\EndWhile
\end{algorithmic}
\end{algorithm}

The full reverse algorithm can be found in Algorithm \ref{alg:reverse}. The algorithm first splits by new line characters. To combine vertically written characters, the algorithm appends them to the position in an original text line. An original text line is determined by those lines which have more than single characters. Note, the algorithm cannot just take the top line as the only text line as the width constraint in VertAttack adds vertical lines throughout the text.

\section{Analysis of Percentage of Words Perturbed}
\begin{figure}
    \centering
    \includegraphics[width=\columnwidth]{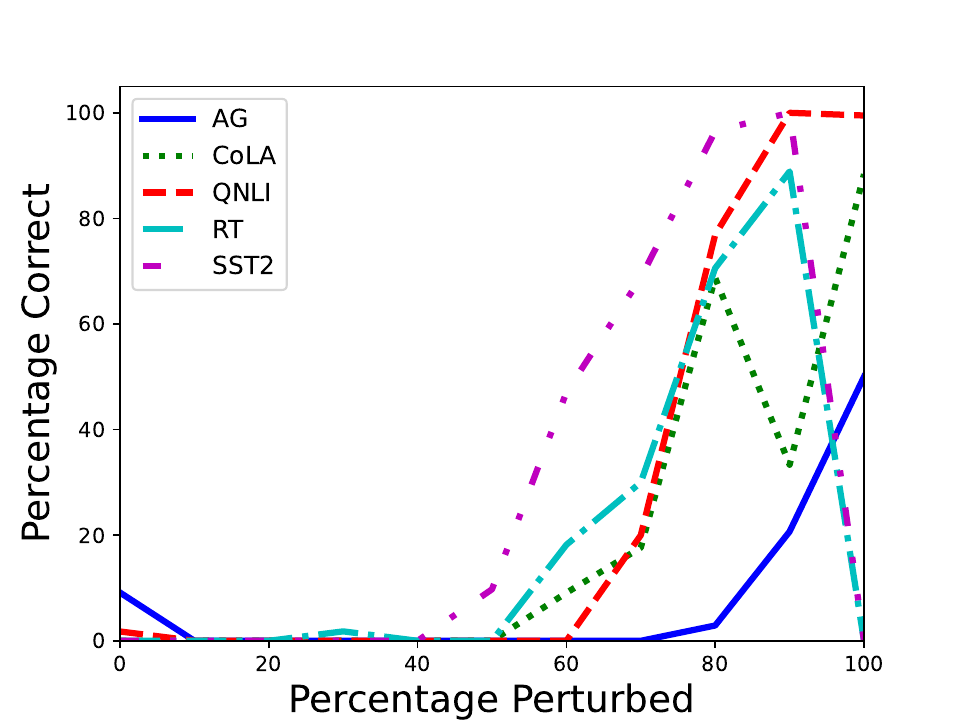}
    \caption{The classifiers' ability to correctly classify text as the amount of words perturbed increases. The classifier examined is BERT, when \textit{VertAttack} uses BERT for feedback.}
    \label{fig:bertcorrectpert}
\end{figure}

For additional understanding of VertAttack, we seek to analyze how the number of words modified by \textit{VertAttack} affects the classifiers. One might postulate that as \textit{VertAttack} modifies more words the classifier does worse, as more and more of the original text is lost. However, through our analysis we find the opposite to be true. 

Figure \ref{fig:bertcorrectpert} graphs BERT's classification ability versus percentage of text perturbed across the 5 examined datasets. Surprisingly, we see that as the percentage of words perturbed increases, the classifier is better equipped to make a correct classification. This may partially be due to a limitation with \textit{VertAttack} compared to some other attacks. Other attacks are able to bring in new words whose embeddings can cause additional confusion for the classifier, but \textit{VertAttack} does not.

\end{document}